*Article*

# Advancing Pancreatic Cancer Prediction with a Next Visit Token Prediction Head on top of Med-BERT

Jianping He [1][+], Laila Rasmy [1][+], Degui Zhi [1],[*] and Cui Tao [2],[*]

1. McWilliams School of Biomedical Informatics, UTHealth at Houston, Houston, TX, USA
2. Department of Artificial Intelligence and Informatics, Mayo Clinic, Jacksonville, FL, USA
+ contributed equally
* Correspondence: degui.zhi@uth.tmc.edu; Tao.Cui@mayo.edu;

**Simple Summary:** Pancreatic cancer (PaCa) is estimated to remain the fourth leading cause of cancer deaths in men, following lung, colon, and prostate cancers, and the third leading cause in women, following lung and breast cancers. PaCa is often referred to as a "silent killer" because symptoms typically manifest only in the late stages of the disease. Consequently, early detection is crucial for improving patient outcomes. This study explores the use of electronic health records (EHRs) to enhance the prediction of PaCa onset. Our research leverages Med-BERT, a foundation model designed for EHR data, to improve early detection using deep learning techniques. By aligning the prediction task with Med-BERT's pretraining task, we aim to enhance its accuracy, especially in scenarios with limited data. This approach can facilitate the earlier detection of PaCa in patients, thereby improving their prognosis.

**Abstract:** Background: Electronic Health Records (EHRs) encompass valuable data essential for disease prediction. The application of artificial intelligence (AI), particularly deep learning, significantly enhances disease prediction by analyzing extensive EHR datasets to identify hidden patterns, facilitating early detection. Recently, numerous foundation models pretrained on extensive data have demonstrated efficacy in disease prediction using EHRs. However, there remains some unanswered questions on how to best utilize such models especially with very small fine-tuning cohorts. Methods: We utilized Med-BERT, an EHR-specific foundation model, and reformulated the disease binary prediction task into a token prediction task and a next visit mask token prediction task to align with Med-BERT's pretraining task format in order to improve the accuracy of pancreatic cancer (PaCa) prediction in both few-shot and fully supervised settings. Results: The reformulation of the task into a token prediction task, referred to as Med-BERT-Sum, demonstrates slightly superior performance in both few-shot scenarios and larger data samples. Furthermore, reformulating the prediction task as a Next Visit Mask Token Prediction task (Med-BERT-Mask) significantly outperforms the conventional Binary Classification (BC) prediction task (Med-BERT-BC) by 3% to 7% in few-shot scenarios with data sizes ranging from 10 to 500 samples. These findings highlight that aligning the downstream task with Med-BERT's pretraining objectives substantially enhances the model's predictive capabilities, thereby improving its effectiveness in predicting both rare and common diseases. Conclusion: Reformatting disease prediction tasks to align with the pretraining of foundation models enhances prediction accuracy, leading to earlier detection and timely intervention. This approach improves treatment effectiveness, survival rates, and overall patient outcomes for PaCa and potentially other cancers.

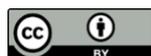

## 1. Introduction

Pancreatic cancer (PaCa) is estimated to remain the fourth leading cause of cancer deaths in men, following lung, colon, and prostate cancers, and the third leading cause in women, following lung and breast cancers [1]. Patients diagnosed in the early stages of PaCa have the potential for a cure through a combination of surgery, chemotherapy, and radiotherapy. However, detecting PaCa at an early stage is highly challenging, as it is typically identified in its later stages [2]. Therefore, it would be significant if PaCa could be detected at the early disease stage to improve prognosis and increase the chances of successful treatment.

Electronic health records (EHRs) data contains a large amount of potentially valuable information for medical research and diagnosis [3]. EHR-based disease prediction can be a very significant aspect of healthcare as it allows for the early detection of illnesses. [4]. In the context of PaCa, emerging studies focus on predicting disease risk using real-world clinical records from large patient cohorts. These studies aim to facilitate the design of cost-effective surveillance programs for early detection [5] [6].

Artificial intelligence (AI) techniques, including machine learning and other computational methods, are increasingly used in PaCa prediction to analyze large amounts of heterogeneous EHR data [2,5–10]. In recent years, transformer-based models have demonstrated significant potential in PaCa prediction. These models can achieve exceptionally high-performance levels when provided with sufficient high-quality training data [2,5–10]. The necessity of large, annotated datasets is particularly critical for transformer-based models, as they require extensive data to learn complex representations of the input domain. In the absence of enough and adequate training data, these models may experience underfitting, leading to suboptimal performance on new, unseen data [11]. Therefore, developing transformer-based models that can effectively perform PaCa prediction in few-shot scenarios is of paramount importance.

A foundation model is a large-scale deep learning model trained on extensive, diverse datasets, capable of performing a wide range of tasks across different domains [12]. These models are highly adaptable, allowing for customization to specific applications without the need for developing new AI systems from scratch. Foundation models serve as a versatile base for creating more specialized models efficiently and cost-effectively, exemplified by AI technologies like BERT, GPT, and others that can understand language, generate text and images, and engage in natural conversations [12]. There are a few specialized foundation models available using structured EHR data and it is crucial to optimize their application to achieve better outcomes. Emphasis should be placed on enhancing the utilization of existing models rather than concentrating on the development of new ones.

Med-BERT is an encoder-based EHR foundation model, i.e., a pre-trained contextualized embedding model specific to disease prediction, which is trained on structured diagnosis codes, medication codes, and procedure codes [4,13]. Although Med-BERT has shown promising performance in disease prediction studies, including pancreatic cancer prediction, there is still potential for further improvement especially in few shot settings [4,13].

We hypothesize that reformulating the downstream task of predicting the onset of PaCa to align with the formulation of the token prediction used for Med-BERT pre-training could enhance its performance, particularly in the few-shot scenario. Our results indicate that this reformulation enhances Med-BERT's performance nearly in all settings, with more pronounced improvements observed in the few-shot scenarios.

## 2. Related Work

Numerous BERT variants, such as BioBERT [14] and ClinicalBERT [15], have been pretrained on clinical texts and are utilized for various clinical NLP tasks. These variations inherit the BERT architecture and pretraining tasks, specifically employing Masked language Model (MLM) objective during the pretraining stage to train most transformer based language models. Prompt tuning incorporates a prompt template—a text segment with mask tokens—into the initial input, transforming various downstream NLP tasks into MLM formats. This adjustment aligns target tasks with the pre-training structure, enhancing compatibility with the foundation models learned attentions within all BERT variants [16].

Studies have shown promising results for prompt tuning in the clinical NLP field. For instance, Peng et al. [17] and He et al. [18] demonstrated its effectiveness in named entity and relation extraction. Taylor et al. [19] successfully applied prompt tuning to several clinical tasks, including ICD-9, ICD-9 Triage Task, In-Hospital Mortality, and Length of Stay in the ICU. Zhang et al. [20] utilized prompt tuning for the classification of clinical notes, while Maharjan et al. [21] applied it to clinical question answering. In all the above studies, the success of the research relies on text-based input for foundation models, where the prediction tasks are converted to a MLM format.

Given the limited availability of BERT-based foundation models trained on structured EHR data, Med-BERT stands out as a model trained on the largest patient cohort, with the full trajectory of over 20 Million patients data [4]. Med-BERT leveraged the original BERT framework, including its architecture and training methodology, and pretrained on structured EHR data. It employed the MLM as one of the pretraining tasks, directly derived from the original BERT paper. This task involved predicting the presence of any code based on its context, with an 80% probability of replacing a code with [MASK], a 10% probability of replacing it with a random code, and a 10% probability of leaving it unchanged. This task is central for the training of any encoder only based model [3].

Therefore, we hypothesized that if the task of PaCa is reformulated similarly to the MLM masked token prediction, it could lead to improved performance. Therefore, we reformulated the PaCa task by 2 key steps. The first is transforming the main outcome objective from a simple binary label for Pancreatic Cancer (Yes/ No) into the probability of the patient getting any Pancreatic Cancer relevant ICD-10 codes in a later visit, being a part of the vocabulary that Med-BERT was pre-trained on. The second, is adding the Masked token to the first future (next) visit, so that the downstream task formulation will be shifted from a simple classification task into MLM task. To our knowledge, no prior studies have attempted to reformulate the disease prediction downstream task in this way to improve EHR based foundation models' performance.

## 3. Materials and Methods

### 3.1. Materials

To evaluate the added value of reformulating the disease prediction downstream task to MLM task to improve prediction accuracy, we utilized the same pancreatic cancer cohort used in the Med-BERT v2 study [13]. This cohort consists of 31,243 patients, including 12,273 cases and 18,970 controls. The dataset was split into training, validation, and test sets in a ratio of 7:1:2, resulting in up to 21,871 patients for fine-tuning, 3,124 patients for validation, and 6,248 patients for testing.

### 3.2. Methods

3.2.1. Reformulating PaCa Onset Task into Masked Language Model (LM)

As illustrated in Figure 1, visits 1, 2, up to visit i correspond to the patient's diagnosis, medications, and procedures until the current visit i. Comonly, the PaCa risk prediction task is approached as a binary classification problem. In contrast, our proposed approach is to estimate the probability of a pancreatic cancer ICD-10 code filling the position of a MASK token added to the next visit (i+1). This reformulates the PaCa prediction task into a MLM task. This prediction task reformulation was technically achievable through two main modifications. The first is the design of the outcome label. So instead of predicting directly if the patient will develop PaCa in the future (simple binary label: Yes/ No), we will predict if the patient will have any of the eight tokens representing the following ICD-10 codes for pancreatic cancer in any future visit, namely {C25.0 Malignant neoplasm of head of pancreas; C25.1 Malignant neoplasm of body of pancreas; C25.2 Malignant neoplasm of tail of pancreas; C25.3 Malignant neoplasm of pancreatic duct; C25.4 Malignant neoplasm of endocrine pancreas; C25.7 Malignant neoplasm of other parts of pancreas; C25.8 Malignant neoplasm of overlapping sites of pancreas; C25.9 Malignant neoplasm of pancreas, unspecified}. This modification transforms the predicted label to utilize a subset of the (tokens) labels used during Med-BERT pretraining phase. Therefore, we evaluated the value added of this modification in two settings, the first using the full patient context vector, the same as we use as the input for the binary classification head, and we further refer to as Token Prediction (MedBERT-Sum). The second is using the learned vector of the masked token at the next visit (i+1) which we refer to as Next visit Mask Token prediction (MedBERT_Mask). It is worth noting here, that unlike the Med-BERT MLM pretraining task, we only used the token embedding for the above ICD-10 codes to calculate only the probabilities of these tokens to fill the Mask position instead of the full tokens list in the vocabulary given our specific PaCa prediction use case, and accordingly our objective here was slightly modified to find if any of the PaCa ICD-10 corresponding tokens have a high probability to fill the Mask position rather than finding the code with the highest chance of being the best fit which was used for the more generalizable pretraining objective.

The second modification is adding the Masked token to the input sequence along with the next future visit (i+1) position, so that the downstream task formulation will be shifted from a simple classification task into MLM task. In this setting, we use the dot product of the contextual vector at the Mask token position with the PaCa label tensor that include the corresponding token embeddings of the eight ICD-10 codes listed above, to estimate the probability of each of the tokens to fill the Mask position, same as described above.

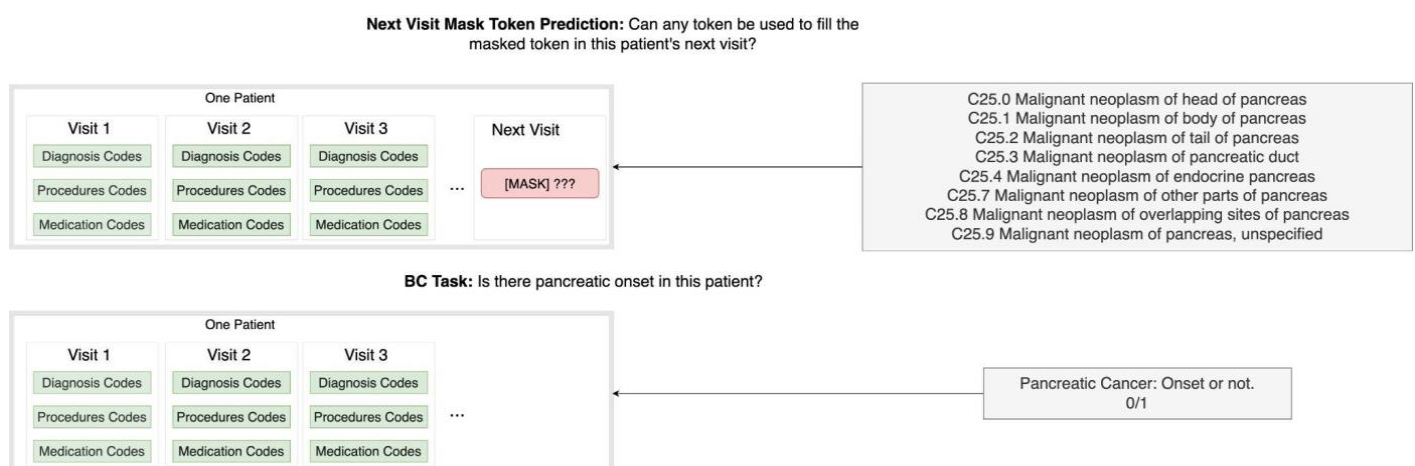

**Figure 1.** The Principle of Converting the PaCa OnSet Task into a Masked LM.

Therefore, our intermediate output is a vector whose length matches the number of PaCa ICD-10 codes (sub-labels), eight in our case. Each element in this vector indicates the probability of this token to present in the next visit. We then use the maximum probability from this intermediate output for the final pancreatic cancer prediction evaluation.

3.2.2. The Scheme of Utilizing Med-BERT for PaCa Onset Prediction

Figure 2 illustrates the scheme of utilizing Med-BERT for PaCa onset prediction. As illustrated in Figure 2, We used the full patient trajectory before the index date (visit i) in a sequential format where clinical codes including diagnosis, medication, and procedures are sorted by the event date and time within visits' sequences. We first mapped the clinical codes to the Med-BERT tokens. Then we project the data to the Med-BERT v2 model. Within Med-BERT, the tokens get projected to the token embedding layer and the visits sequence to the visit embedding layer to generate the static patient representation, which is then projected to the transformer layers to get the contextualized embeddings for each patient.

In the original PaCa onset prediction task, we initially derived patient contextualized representations with dimensions of [sequence length, embedding dimension]. Subsequently, by summing the values of the contextualized embeddings along the sequence length dimension, we generated a patient contextual vector with dimensions of [1, embedding dimension]. This patient contextual vector was then employed in the token prediction head as explained above in 3.2.1 as well as the model's binary classification prediction head to perform the PaCa onset prediction task. This approach is referred to as Med-BERT-BC in this paper.

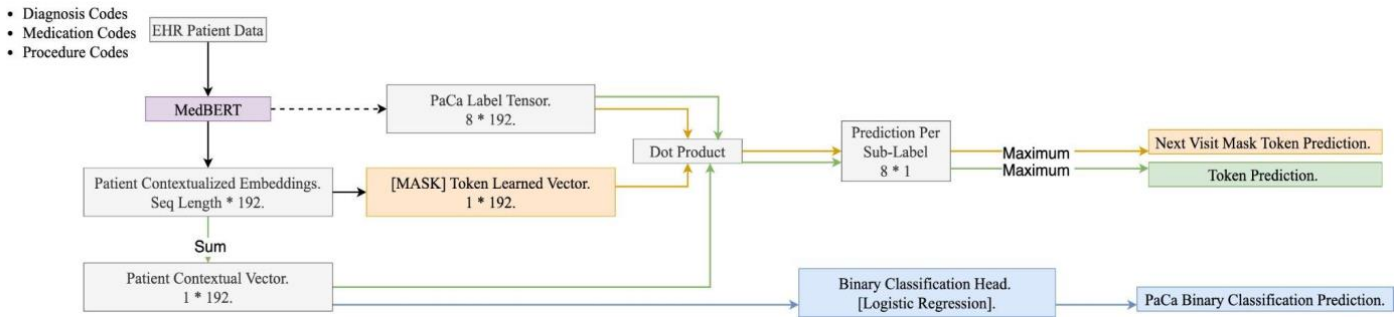

**Figure 2.** The Scheme of Utilizing Med-BERT for PaCa Onset Prediction.

By reformulating the Pancreatic Cancer (PaCa) onset task into a next visit mask token prediction task, our objective is to identify the most probable token that can replace the masked token from the subset of the target concept 'PaCa'. To achieve this, we extracted the PaCa Label Tensor from the Med-BERT token embedding layer, with a dimension of [8, embedding dimension]. We multiplied the masked token learned vector by the PaCa Label Tensor, resulting in a vector whose length corresponds to the number of PaCa labels. The highest score within this vector was selected to determine if a PaCa-related label could be assigned to the masked token. This approach is referred to as Med-BERT-Mask. In addition, by reformulating the PaCa onset task into a token prediction task, we performed a dot product between the patient contextual vector (utilized in the binary classification prediction head to calculate the probability of PaCa onset) and the PaCa Label Tensor. This approach shares similarities with Med-BERT-Mask and is referred to as Med-BERT-Sum in this paper. In Figure 2, the blue path represents the binary prediction, Med-BERT-BC; the orange path denotes the next visit mask token prediction, Med-BERT-Mask; and the green path signifies the token prediction, Med-BERT-Sum.

### 3.2.3. Baseline Models

In our study, the baseline models employed include Logistic Regression (LR), Long Short-Term Memory (LSTM), Bidirectional LSTM (BiLSTM), Gated Recurrent Unit (GRU), and Bidirectional GRU (BiGRU). These models were chosen due to their proven effectiveness in handling sequential data, allowing us to benchmark our proposed approach against well-established architectures in the field [4].

### 3.2.4. Experimental Settings

We evaluated the model's performance using the Area Under the Receiver Operating Characteristic Curve (AUROC). The sequence length was fixed at 64, truncating any sequences that exceeded this length. Clinical diagnoses, procedures, and medications for each patient were arranged in reverse chronological order. By setting the sequence length to 64, we retained the most recent 64 codes up to the patient's last visit. The batch size was configured to 100. For the learning rate, we employed a value of 1e-3 for the Med-BERT models, GRU, and LSTM, while a learning rate of 1e-5 was used for BiGRU and BiLSTM.

We assessed the performance of all models in both few-shot scenarios and fully supervised settings. The data sample sizes ranged in [10, 20, 30, 40, 50, 100, 200, 300, 400, 500, 1000, 2000, 3000, 4000, 5000, 100000], as well as the full dataset. For data sizes of 10, 20, 30, 40, 50, 100, 200, 300, 400, 500, and 1000, we used an equal split of positive and negative instances for both the training and validation datasets. For example, with a data size of 10, we utilized 5 positive instances and 5 negative instances for both fine-tuning and validation. For data sizes of 2000, 3000, 4000, 5000, and 100,000, we employed the entire validation dataset for validation, while maintaining an equal split of positive and negative instances for the fine-tuning dataset. For the full data size, the entire fine-tuning and validation datasets were used. Regardless of the experimental data size, the complete test dataset was used for testing. For each data size, we conducted 3 runs to calculate the mean and standard deviation of the AUROC.

## 4. Results

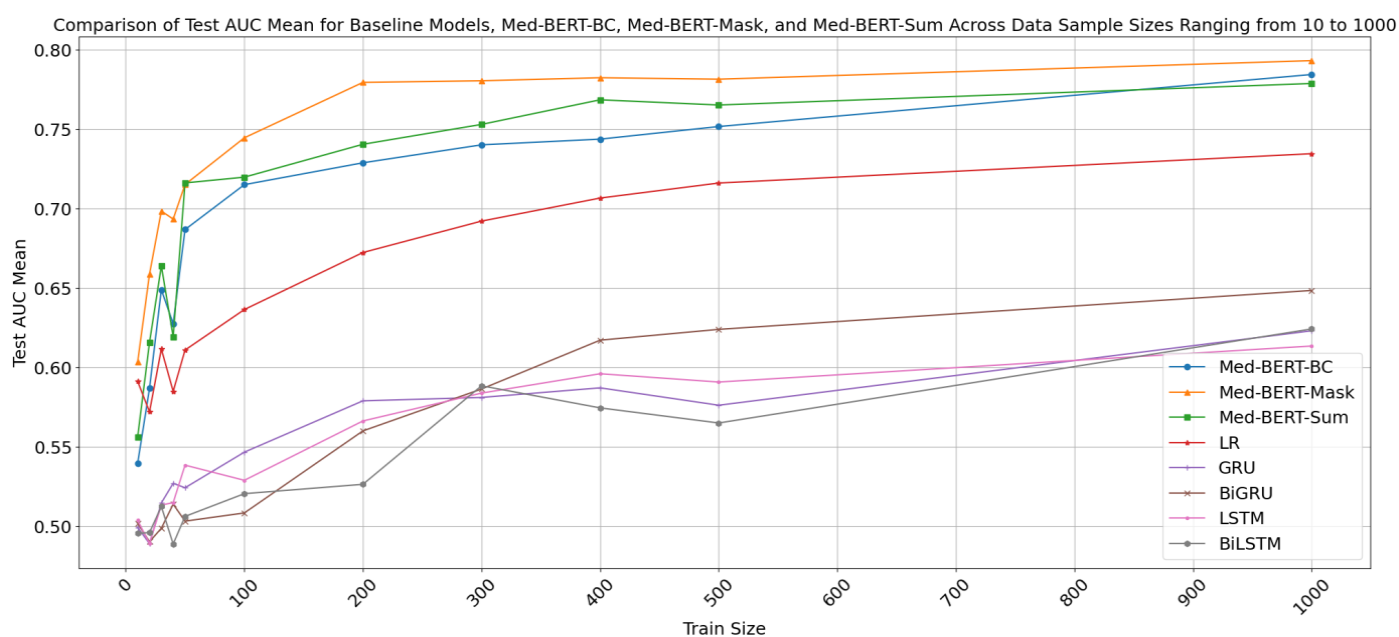

**Figure 3.** Comparison of Test AUC Mean for Baseline Models, Med-BERT-BC, Med-BERT-Mask, and Med-BERT-Sum Across Data Sample Sizes Ranging from 10 to 1000.

Figures 3, and 4 illustrate the AUC mean on the test set for Baseline Models, Med-BERT-BC, Med-BERT-Mask, and Med-BERT-Sum across data sample sizes ranging from 10 to 1000, and 1000 to 10000, respectively. The x-axis represents the training size, while the y-axis shows the AUC mean on the test set. From Figure 3 and Figure 4, it is evident that all models exhibit an increase in performance as the fine-tuning data size grows, indicating improved performance with more data. Notably, Med-BERT models, including Med-BERT-BC, Med-BERT-Sum and Med-BERT-Mask, consistently achieve significantly higher AUC scores compared to the baseline models, highlighting their superior performance.

As illustrated in Figure 3 and Figure 4, Med-BERT-Sum demonstrates a modest improvement in average AUC performance compared to Med-BERT-BC across almost all data sizes, highlighting the effectiveness of leveraging Med-BERT pretraining vocabulary token embeddings for the PaCa label tensor. Compared to Med-BERT-Sum, Med-BERT-Mask demonstrates a significant improvement in performance at very low fine-tuning sample sizes (10-500 samples) as shown in Supplementary Table 1. Furthermore, when compared to Med-BERT-BC, Med-BERT-Mask exhibits a marked increase in performance, with average AUC improvements ranging from 3% to 7% in few-shot scenarios with data sample sizes from 10 to 500 (Supplementary Table 1). This suggests that reformulating the prediction task into a MLM task enhances performance in few-shot settings.

As the training size increases, the performance of all three Med-BERT models stabilizes. When the data size is below 1000, Med-BERT-Mask continues to outperform the other models, achieving approximately 0.79 AUC with 1000 samples, which is 1% higher than Med-BERT-BC's 0.78 AUC. When the data size is larger than 1000, Med-BERT-BC and Med-BERT-Sum both exhibit performance improvements with increasing training sizes, with Med-BERT-BC reaching an AUC of approximately 0.82 at 10,000 samples. Med-BERT-Mask initially shows a decline in performance after 1000 samples, followed by a gradual improvement, but does not surpass the performance of Med-BERT-BC and Med-BERT-Sum beyond 1000 samples. Med-BERT-Sum achieves the highest AUC of approximately 0.83 with 10,000 training samples, consistently outperforming Med-BERT-Mask once the fine-tuning sample size exceeds 1000. Additionally, Med-BERT-Sum slightly surpasses Med-BERT-BC as training sizes increase, and consistently demonstrates superior performance compared to Med-BERT-BC across training sizes ranging from 2000 to the full fine-tuning set of 21,871 samples. These observations suggest that with sufficient training data, Med-BERT-BC performs well, while utilizing Med-BERT pretraining vocabulary token embeddings for the PaCa label tensor is advantageous in both few-shot scenarios and larger training cohorts.

The observed sharp decrease in AUC mean for training sizes between 40 and 50, as shown in Figure 3, may be attributed to the presence of outliers and the inherent variability in smaller datasets. To further validate the model's robustness, we conducted additional repeated experiments on smaller datasets with training sizes of less than 100, performing 10 runs to assess the consistency of the results. These findings are presented in Supplementary Figure 6. The results indicate that while outliers can occasionally impact performance in smaller datasets, the overall trend remains consistent. Notably, Med-BERT-Mask demonstrates robust performance with limited data, reinforcing the model's effectiveness under these conditions.

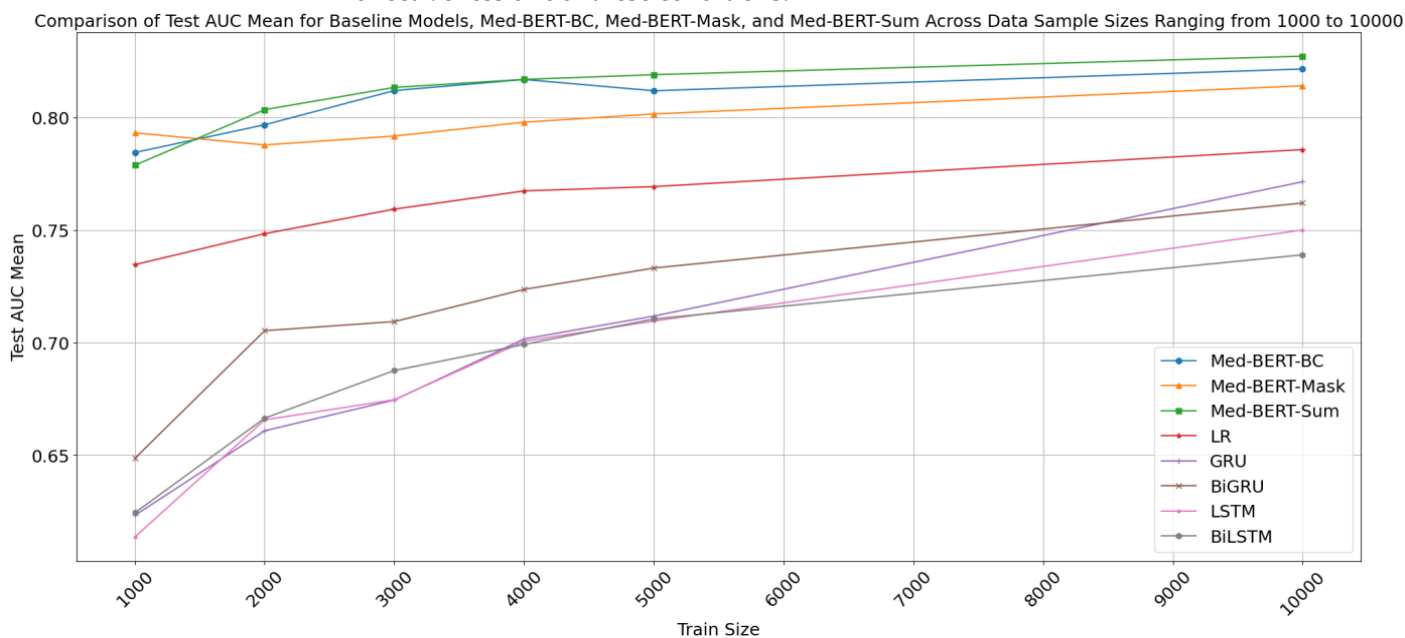

**Figure 4.** Comparison of Test AUC Mean for Baseline Models, Med-BERT-BC, Med-BERT-Mask, and Med-BERT-Sum Across Data Sample Sizes Ranging from 1000 to 10000.

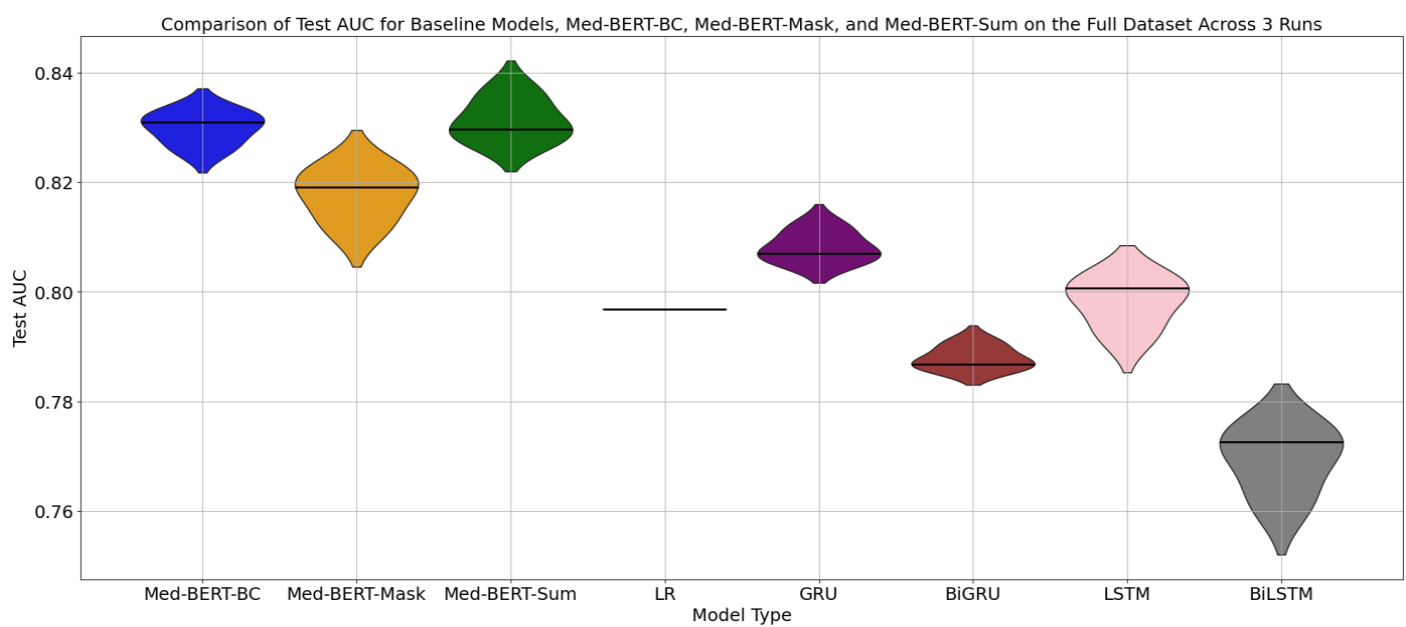

**Figure 5.** Comparison of Test AUC for Baseline Models, Med-BERT-BC, Med-BERT-Mask, and Med-BERT-Sum on the Full Dataset Across 3 Runs.

Figure 5 is a violin plot comparing the Test AUC for eight different models: Med-BERT-BC, Med-BERT-Mask, Med-BERT-Sum, LR, GRU, BiGRU, LSTM and BiLSTM across 3 runs using the full dataset. The x-axis represents the model type, and the y-axis shows the Test AUC scores. Each "violin" represents the distribution of the Test AUC values for one model type. The solid lines within each plot indicate the median. We can see that with fully supervised settings, Med-BERT models generally could achieve better performance than baseline models.

Med-BERT-BC has a relatively narrow distribution, indicating a consistent performance across the 3 runs. Med-BERT-Mask has a wider distribution, indicating more variability in performance. The Test AUC values for this model are spread out more compared to Med-BERT-BC. Similar to Med-BERT-Mask, Med-BERT-Sum also shows a wider distribution of Test AUC values, indicating variability in the results. While Med-BERT-Sum' results lack stability, it demonstrates the potential to achieve higher performance in certain runs.

Supplementary Table 1 presents the detailed results including the AUC mean, standard deviation, and performance boost compared to Med-BERT-BC on the test set for all models across all data sample sizes. The results indicate that Med-BERT-Mask significantly outperforms Med-BERT-BC when trained on smaller datasets, highlighting the advantage of reformulating the downstream task into the MLM task in few-shot scenarios. Additionally, Med-BERT-Sum demonstrates a slight performance improvement over Med-BERT-BC across both few-shot and larger dataset settings. These findings suggest that reformulating the PaCa label tensor to utilize Med-BERT's pretraining vocabulary token embeddings consistently enhances model performance. In contrast, the MLM task formulation proves particularly beneficial in few-shot settings, providing substantial performance gains under these conditions.

## 5. Discussion

Med-BERT-BC utilizes binary classification directly on the pancreatic cancer prediction task. Med-BERT-Sum uses the Med-BERT pretraining vocabulary token embeddings for the PaCa label tensor, which is the primary distinction from Med-BERT-BC. Med-BERT-Sum demonstrates slight superior performance compared to Med-BERT-BC across most data sizes (ranging from 10 to fully supervised setting), suggesting that utilizing the Med-BERT pretraining vocabulary token embeddings for the PaCa label tensor is effective for boosting Med-BERT. Med-BERT-Mask not only incorporates the Med-BERT pretraining vocabulary token embeddings but also reformulates the binary classification task into a MLM task, aligning with Med-BERT's pretraining objective. The observation that Med-BERT-Mask achieves significantly higher AUC scores compared to Med-BERT-BC with limited fine-tuning data underscores the efficacy of reformulating the PaCa prediction downstream task into a MLM framework as hypothesized. These findings collectively demonstrate that reformulating the PaCa prediction task into the formulation of the token prediction used for Med-BERT pre-training could enhance its predictive capabilities for pancreatic cancer even with the availability of few tens of cases data and equal numbers of controls. So for example with a 100 PaCa cases and 100 controls, we can fine-tune Med-BERT-Mask for PaCa prediction and achieve prediction accuracy of around 78%, which is similar to the AUC that can be achieved after fine-tuning a binary classification head on top of Med-BERT for 1000 samples. Consequently, this study encourages us to formulate disease prediction downstream tasks in a manner akin to the pretraining task format when utilizing foundation models like Med-BERT in order to maximize their performance in few shot settings.

Med-BERT-Sum slightly outperforms Med-BERT-BC across all the data sizes, demonstrating the added value of using the pretraining vocabulary tokens as an intermediate label to predict. Med-BERT-Mask excels in few-shot scenarios due to its ability to effectively leverage pre-trained knowledge through task reformulation. This approach optimally utilizes the generalized patterns and representations from the pre-trained language model, enhancing performance with limited fine-tuning data. However, as the fine-tuning dataset size increases, the classification head can better converge in Med-BERT-BC, which allows the Med-BERT to learn more discriminative features tailored to the specific task of predicting PaCa onset. Therefore, the extensive task-specific data reduces the need for reformulation, as direct optimization yields better and stable results.

In the domain of clinical decision support, firstly, our method enhances the accuracy of PaCa prediction. By offering more reliable predictions, our approach can aid healthcare professionals in making better-informed decisions, ultimately leading to improved patient outcomes. This provides substantial benefits for the clinical field, including PaCa, other cancers, and even non-cancerous diseases. Furthermore, the challenge of diagnosing rare diseases is exacerbated by the limited datasets available for training models to achieve high performance with new patients. Our method's strong performance in few-shot scenarios demonstrates the feasibility of AI-based clinical decision support systems for rare diseases. This success addresses the unique challenge posed by infrequent diseases and encourages further innovation in AI-driven healthcare, potentially extending benefits to a broader range of medical applications.

If the clinical workflow incorporates BERT-based models like Med-BERT, our approach suggests reformulating the downstream task into a token prediction task (Med-BERT-Sum) or a masked token prediction task (Med-BERT-Mask) based on the availability of training data. Med-BERT-Mask is advantageous in scenarios with limited data, while Med-BERT-Sum performs better with larger datasets. If clinical workflows employ models that are not BERT-based, the downstream task can still be tailored to fit the pretraining task format of the specific model in use. This strategy involves analyzing the pretraining objective of the foundation model and reformulating the clinical prediction task to resemble this objective, which can significantly enhance the model's performance. However, we acknowledge the challenges in implementing this approach, such as the need for expertise in understanding and adapting the task formulations.

Our method presents several limitations that warrant further investigation in future research. First, this study utilized the PaCa Onset Cohort, as employed in the original Med-BERT v2 study. Notably, this cohort is based on claims data, which is retrospective and generated for billing purposes rather than clinical objectives. Consequently, the sensitivity for pancreatic cancer within this dataset is relatively low. Furthermore, the recorded time for PaCa onset often lags behind the true diagnosis due to the inherent difficulty in diagnosing pancreatic cancer, which typically results in a billing entry only after a confirmed diagnosis. This characteristic leads to a high number of true positives, reducing the risk of false positives, and contributing to an elevated positive predictive value (PPV). Regarding the control group, we undertook rigorous steps to minimize false negatives by excluding all other cancer patients, including those with benign tumors, from the cohort. Ideally, validation against manual chart reviews would enhance accuracy; however, this study prioritized algorithm development, making this limitation beyond our current scope. In future work, we plan to utilize cohorts derived from clinical data instead of billing data to mitigate these issues. Additionally, the cohort used in this study provided a relatively balanced dataset for training and evaluation purposes. However, we acknowledge that this balanced distribution does not mirror real-world conditions, where the incidence of pancreatic cancer is significantly lower in the general population. In future research, addressing class imbalance during the cohort inclusion and exclusion stages will be critical. This could involve intentionally constructing an imbalanced cohort to better represent real-world scenarios, thereby enhancing the model's applicability in practice. Moreover, the current evaluation was restricted to the prediction of PaCa onset, which limits our understanding of the method's generalizability to different diseases and clinical contexts. To ensure robustness and broader applicability, future studies should incorporate evaluations across a diverse array of diseases, including various cancers and non-cancerous conditions. Finally, our testing was confined to a single clinical foundation model, Med-BERT. Future evaluations should extend to other clinical foundation models to comprehensively assess the method's performance.

## 6. Conclusions

In this study, we reformulated the downstream task of predicting the onset of PaCa to align with the token prediction used in Med-BERT pre-training. This reformulation enhanced Med-BERT fine-tuning efficacy, resulting in superior performance compared to the conventional BC prediction design in both few-shot and fully supervised settings. Our findings highlight the potential of leveraging foundation models more effectively by aligning downstream tasks with their pretraining objectives. This strategy offers a promising avenue for developing more robust AI clinical support systems for PaCa, other cancers, and non-cancerous diseases, thereby improving patient outcomes. Future research will focus on validating these findings across diverse cancer and non-cancer contexts, as well as on different foundation models, and exploring further optimizations to enhance cancer diagnosis systems..

**Author Contributions:** Conceptualization, C.T., D.Z., and L.R.; methodology, L.R. and J.H.; validation, L.R. and J.H.; formal analysis, L.R. and J.H.; investigation, L.R. and J.H.; resources, L.R. and J.H.; data curation, L.R. and J.H.; writing—original draft preparation, J.H.; writing—review and editing, L.R., C.T. and D.Z.; visualization, L.R. and J.H.; supervision, C.T. and D.Z.; project administration, C.T. and D.Z.; funding acquisition, C.T., D.Z., and L.R.. All authors have read and agreed to the published version of the manuscript.

**Funding:** Research reported in this publication was supported by the National Library of Medicine of the National Institutes of Health under Award Number R01LM014249, National Institute on Aging Awards numbers R01AG083039 and U01AG070112-02S1, and the American Heart Association, grant number 19GPSGC35180031.

The content is solely the responsibility of the authors and does not necessarily represent the official views of the National Institutes of Health.

**Institutional Review Board Statement:** The study was conducted in accordance with the Declaration of Helsinki, and approved by the Committee for the Protection of Human Subjects (UTHSC-H IRB) under protocol HSC-SBMI-13-0549.

**Informed Consent Statement:** Informed consent was obtained from all subjects involved in the study.

**Conflicts of Interest:** The authors declare no conflicts of interest.

**Supplementary Table 1.** The Test AUC Mean, Standard Deviation, and Performance Boost Compared to Med-BERT-BC for All Models across All Data Sample Sizes.

| Train_Size | Function_Type | Test_AUC mean (%) | Performance Boost (%) | Test_AUC std |
|---|---|---|---|---|
| 10 | LR | 59.13 | 5.15 | 0.02 |
| 10 | GRU | 49.95 | -4.03 | 0.03 |
| 10 | BiGRU | 50.21 | -3.77 | 0.02 |
| 10 | LSTM | 50.39 | -3.59 | 0.01 |
| 10 | BiLSTM | 49.57 | -4.41 | 0.02 |
| 10 | Med-BERT-BC | 53.98 | 0.00 | 0.04 |
| 10 | Med-BERT-Sum | 55.64 | 1.66 | 0.05 |
| 10 | **Med-BERT-Mask** | **60.35** | **6.37** | **0.05** |
| 20 | LR | 57.22 | -1.47 | 0.07 |
| 20 | GRU | 48.92 | -9.77 | 0.02 |
| 20 | BiGRU | 49.04 | -9.65 | 0.01 |
| 20 | LSTM | 48.93 | -9.76 | 0.02 |
| 20 | BiLSTM | 49.62 | -9.07 | 0.01 |
| 20 | Med-BERT-BC | 58.69 | 0.00 | 0.10 |
| 20 | Med-BERT-Sum | 61.58 | 2.89 | 0.06 |
| 20 | **Med-BERT-Mask** | **65.88** | **7.19** | **0.06** |
| 30 | LR | 61.17 | -3.73 | 0.03 |
| 30 | GRU | 51.48 | -12.92 | 0.03 |
| 30 | BiGRU | 49.89 | -15.01 | 0.02 |
| 30 | LSTM | 51.35 | -13.55 | 0.03 |
| 30 | BiLSTM | 51.26 | -13.64 | 0.01 |
| 30 | Med-BERT-BC | 64.90 | 0.00 | 0.04 |
| 30 | Med-BERT-Sum | 66.40 | 1.50 | 0.05 |
| 30 | **Med-BERT-Mask** | **69.85** | **4.95** | **0.06** |
| 40 | LR | 58.51 | -4.25 | 0.04 |
| 40 | GRU | 52.72 | -10.04 | 0.01 |
| 40 | BiGRU | 51.41 | -11.35 | 0.00 |
| 40 | LSTM | 51.50 | -11.26 | 0.01 |
| 40 | BiLSTM | 48.90 | -13.86 | 0.02 |
| 40 | Med-BERT-BC | 62.76 | 0.00 | 0.05 |
| 40 | Med-BERT-Sum | 61.94 | -0.82 | 0.03 |
| 40 | **Med-BERT-Mask** | **69.35** | **6.59** | **0.04** |
| 50 | LR | 61.10 | -7.59 | 0.05 |
| 50 | GRU | 52.44 | -16.25 | 0.01 |
| 50 | BiGRU | 50.34 | -18.35 | 0.01 |
| 50 | LSTM | 53.86 | -14.83 | 0.04 |
| 50 | BiLSTM | 50.64 | -18.05 | 0.01 |
| 50 | Med-BERT-BC | 68.69 | 18.05 | 0.04 |
| 50 | **Med-BERT-Sum** | **71.62** | **2.93** | **0.02** |
| 50 | Med-BERT-Mask | 71.52 | 2.83 | 0.05 |
| 100 | LR | 63.65 | -7.86 | 0.02 |
| 100 | GRU | 54.68 | -16.83 | 0.03 |
| 100 | BiGRU | 50.86 | -20.65 | 0.02 |
| 100 | LSTM | 52.91 | -18.60 | 0.02 |
| 100 | BiLSTM | 52.07 | -19.44 | 0.04 |
| 100 | Med-BERT-BC | 71.51 | 0.00 | 0.03 |
| 100 | Med-BERT-Sum | 71.99 | 0.48 | 0.02 |
| 100 | **Med-BERT-Mask** | **74.46** | **2.95** | **0.01** |
| 200 | LR | 67.24 | -5.65 | 0.03 |
| 200 | GRU | 57.91 | -14.98 | 0.02 |
| 200 | BiGRU | 56.02 | -16.87 | 0.04 |
| 200 | LSTM | 56.64 | -16.25 | 0.01 |
| 200 | BiLSTM | 52.66 | -20.23 | 0.03 |
| 200 | Med-BERT-BC | 72.89 | 0.00 | 0.01 |
| 200 | Med-BERT-Sum | 74.05 | 1.16 | 0.01 |
| 200 | **Med-BERT-Mask** | **77.95** | **5.06** | **0.00** |
| 300 | LR | 69.22 | -4.80 | 0.00 |
| 300 | GRU | 58.12 | -15.90 | 0.02 |
| 300 | BiGRU | 58.65 | -15.37 | 0.03 |
| 300 | LSTM | 58.40 | -15.62 | 0.02 |
| 300 | BiLSTM | 58.86 | -15.16 | 0.02 |
| 300 | Med-BERT-BC | 74.02 | 15.16 | 0.00 |
| 300 | Med-BERT-Sum | 75.30 | 1.28 | 0.01 |
| 300 | **Med-BERT-Mask** | **78.05** | **4.03** | **0.01** |
| 400 | LR | 70.67 | -3.71 | 0.01 |
| 400 | GRU | 58.73 | -15.65 | 0.03 |
| 400 | BiGRU | 61.73 | -12.65 | 0.03 |
| 400 | LSTM | 59.61 | -14.77 | 0.01 |
| 400 | BiLSTM | 57.47 | -16.91 | 0.04 |
| 400 | Med-BERT-BC | 74.38 | 16.91 | 0.00 |
| 400 | Med-BERT-Sum | 76.85 | 2.47 | 0.00 |
| 400 | **Med-BERT-Mask** | **78.24** | **3.86** | **0.01** |
| 500 | LR | 71.61 | -3.56 | 0.01 |
| 500 | GRU | 57.63 | -17.54 | 0.02 |
| 500 | BiGRU | 62.41 | -12.76 | 0.01 |
| 500 | LSTM | 59.09 | -16.08 | 0.02 |
| 500 | BiLSTM | 56.51 | -18.66 | 0.07 |
| 500 | Med-BERT-BC | 75.17 | 18.66 | 0.01 |
| 500 | Med-BERT-Sum | 76.52 | 1.35 | 0.01 |
| 500 | **Med-BERT-Mask** | **78.15** | **2.98** | **0.01** |
| 1000 | LR | 73.46 | -4.99 | 0.00 |
| 1000 | GRU | 62.32 | -16.13 | 0.01 |
| 1000 | BiGRU | 64.86 | -13.59 | 0.01 |
| 1000 | LSTM | 61.36 | -17.09 | 0.01 |
| 1000 | BiLSTM | 62.43 | -16.02 | 0.01 |
| 1000 | Med-BERT-BC | 78.45 | 16.02 | 0.01 |
| 1000 | Med-BERT-Sum | 77.88 | -0.57 | 0.01 |
| 1000 | **Med-BERT-Mask** | **79.32** | **0.87** | **0.01** |
| 2000 | LR | 74.83 | -4.85 | 0.01 |
| 2000 | GRU | 66.08 | -13.60 | 0.02 |
| 2000 | BiGRU | 70.53 | -9.15 | 0.01 |
| 2000 | LSTM | 66.56 | -13.12 | 0.00 |
| 2000 | BiLSTM | 66.64 | -13.04 | 0.01 |
| 2000 | Med-BERT-BC | 79.68 | 13.04 | 0.00 |
| 2000 | **Med-BERT-Sum** | **80.35** | **0.67** | **0.01** |
| 2000 | Med-BERT-Mask | 78.78 | -0.90 | 0.00 |
| 3000 | LR | 75.93 | -5.27 | 0.01 |
| 3000 | GRU | 67.45 | -13.75 | 0.01 |
| 3000 | BiGRU | 70.93 | -10.27 | 0.00 |
| 3000 | LSTM | 67.46 | -13.47 | 0.01 |
| 3000 | BiLSTM | 68.76 | -12.44 | 0.01 |
| 3000 | Med-BERT-BC | 81.20 | 12.44 | 0.01 |
| 3000 | **Med-BERT-Sum** | **81.34** | **0.14** | **0.00** |
| 3000 | Med-BERT-Mask | 79.18 | -2.02 | 0.01 |
| 4000 | LR | 76.74 | -4.95 | 0.00 |
| 4000 | GRU | 70.16 | -11.53 | 0.01 |
| 4000 | BiGRU | 72.36 | -9.33 | 0.01 |
| 4000 | LSTM | 70.06 | -11.63 | 0.01 |
| 4000 | BiLSTM | 69.91 | -11.78 | 0.01 |
| 4000 | **Med-BERT-BC** | **81.69** | **0.00** | **0.00** |
| 4000 | **Med-BERT-Sum** | **81.69** | **0.00** | **0.00** |
| 4000 | Med-BERT-Mask | 79.79 | -1.90 | 0.00 |
| 5000 | LR | 76.93 | -4.26 | 0.00 |
| 5000 | GRU | 71.17 | -10.02 | 0.01 |
| 5000 | BiGRU | 73.31 | -7.88 | 0.01 |
| 5000 | LSTM | 70.95 | -10.24 | 0.01 |
| 5000 | BiLSTM | 71.05 | -10.14 | 0.01 |
| 5000 | Med-BERT-BC | 81.19 | 0.00 | 0.00 |
| 5000 | **Med-BERT-Sum** | **81.90** | **0.71** | **0.00** |
| 5000 | Med-BERT-Mask | 80.16 | -1.03 | 0.01 |
| 10000 | LR | 78.57 | -3.59 | 0.00 |
| 10000 | GRU | 77.15 | -5.01 | 0.01 |
| 10000 | BiGRU | 76.20 | -5.96 | 0.00 |
| 10000 | LSTM | 75.00 | -7.16 | 0.01 |
| 10000 | BiLSTM | 73.90 | -8.26 | 0.01 |
| 10000 | Med-BERT-BC | 82.16 | 8.26 | 0.00 |
| 10000 | **Med-BERT-Sum** | **82.73** | **0.57** | **0.00** |
| 10000 | Med-BERT-Mask | 81.41 | -0.75 | 0.00 |
| Full Data Size (21871) | LR | 79.69 | -3.31 | 0.00 |
| Full Data Size (21871) | GRU | 80.82 | -2.18 | 0.00 |
| Full Data Size (21871) | BiGRU | 78.79 | -4.21 | 0.00 |
| Full Data Size (21871) | LSTM | 79.82 | -3.18 | 0.00 |
| Full Data Size (21871) | BiLSTM | 76.93 | -6.07 | 0.01 |
| Full Data Size (21871) | Med-BERT-BC | 83.00 | 0.00 | 0.00 |
| Full Data Size (21871) | **Med-BERT-Sum** | **83.13** | **0.13** | **0.00** |
| Full Data Size (21871) | Med-BERT-Mask | 81.78 | -1.22 | 0.00 |

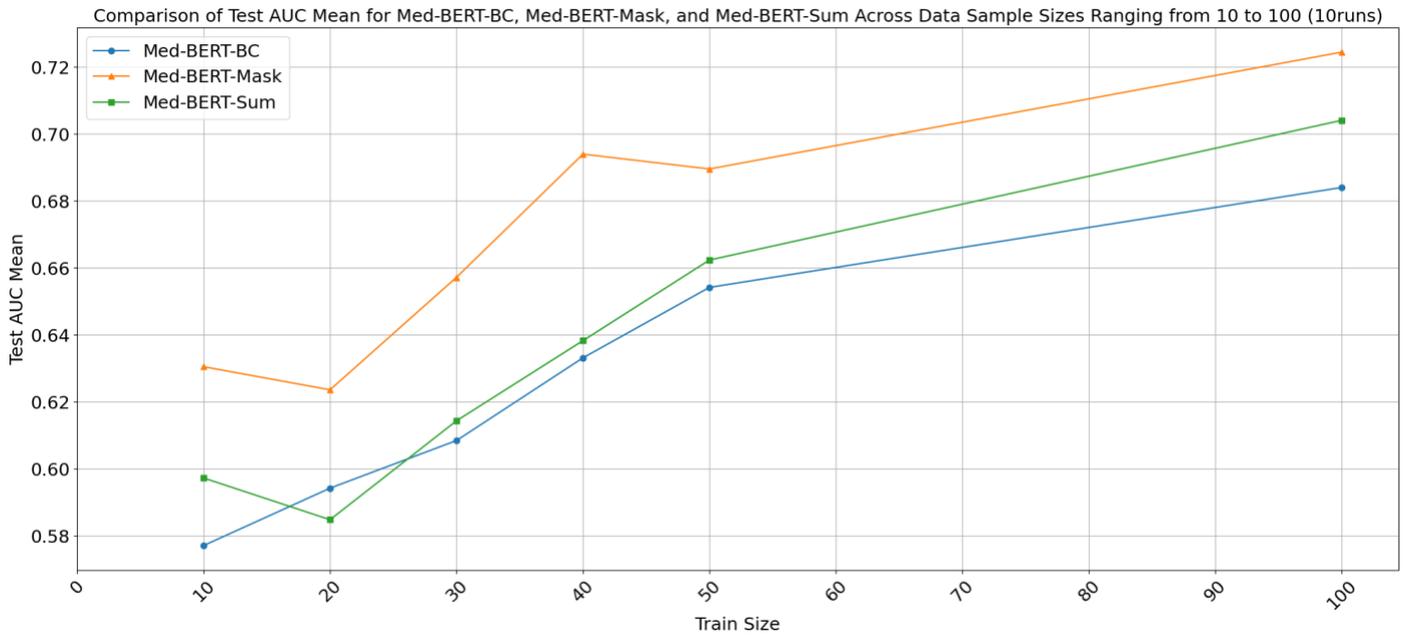

**Figure 6.** Comparison of Test AUC Mean for Med-BERT-BC, Med-BERT-Mask, and Med-BERT-Sum Across Data Sample Sizes Ranging from 1000 to 100 (10 runs).